# Magic Inference Rules for Probabilistic Deduction under Taxonomic Knowledge

**Thomas Lukasiewicz**
Institut für Informatik, Universität Gießen
Arndtstr. 2, D-35392 Gießen, Germany
E-mail: lukasiewicz@informatik.uni-giessen.de

## Abstract

We present locally complete inference rules for probabilistic deduction from taxonomic and probabilistic knowledge-bases over conjunctive events. Crucially, in contrast to similar inference rules in the literature, our inference rules are locally complete for conjunctive events and under additional taxonomic knowledge. We discover that our inference rules are extremely complex and that it is at first glance not clear at all where the deduced tightest bounds come from. Moreover, analyzing the global completeness of our inference rules, we find examples of globally very incomplete probabilistic deductions. More generally, we even show that all systems of inference rules for taxonomic and probabilistic knowledge-bases over conjunctive events are globally incomplete. We conclude that probabilistic deduction by the iterative application of inference rules on interval restrictions for conditional probabilities, even though considered very promising in the literature so far, seems very limited in its field of application.

## 1 INTRODUCTION

Representing and reasoning with uncertain knowledge has gained growing importance in the recent decades. The literature contains many different formalisms and methodologies for tackling uncertainty. Most of them are directly or indirectly based on probability theory.

In this paper, we focus on interval restrictions for conditional probabilities as probabilistic knowledge. The considered probabilistic deduction problems consist of a probabilistic knowledge-base and a probabilistic query. We give a classical example. As a probabilistic knowledge-base, we take the probabilistic knowledge that all ostriches are birds, that ostriches do not fly, that at least 95% of all birds fly, and that not more than 10% of all birds are ostriches. As a probabilistic query, we may wonder about the entailed greatest lower bound and the entailed least upper bound for the rate of all birds that are ostriches. The solution to this probabilistic deduction problem is 0% for the entailed greatest lower bound, and 5% for the entailed least upper bound.

This kind of probabilistic deduction problems can be solved in a global approach by linear programming or in a local approach by the iterative application of inference rules. The global approach by linear programming (see, for example, [21], [12], [22], [10], [15], [14], [3], and [18]) can be performed within rich probabilistic languages capable of representing many facets of probabilistic knowledge (see especially [10]). Probabilistic deduction by linear programming is globally complete, that is, it really produces the requested tightest bounds entailed by the whole probabilistic knowledge-base. However, it generally runs in exponential time in the size of the probabilistic deduction problems. Moreover, it cannot provide any explanatory informations on how the deduced results are obtained.

Mainly to overcome these deficiencies, researchers started to work on local techniques based on inference rules. The local approach (see, for example, [7], [9], [2], [8], [25], [11], [13], and [16]) is generally performed within more restricted probabilistic languages. The iterative application of inference rules is very rarely and only within very restricted probabilistic languages globally complete (see [11] for an example of globally complete local probabilistic deduction in a very restricted framework). Moreover, if the inference rules allow complex events, then they are generally even not locally complete anymore, that is, they generally even do not produce the tightest bounds entailed by the partial probabilistic knowledge in their premises (see [11] and [13] for inference rules that are locally complete only for complex events that are not logically related). Local approaches are generally expected to be more efficient than global ones. Furthermore, they can elucidate the deduction process by the sequence of applied inference rules.

The local approach has been considered very promising in the literature so far. However, its major drawback for prac-tical applications is its global incompleteness. In particular, it is very disappointing that even the inference rules are generally not locally complete anymore for complex events. Hence, the first motivating idea of this paper is to elaborate new inference rules that are locally complete for complex events. Following this idea, we also hope to make a big step towards global completeness.

Coming back to our introductory example, we observe that the sentences that all ostriches are birds and that ostriches do not fly are not purely probabilistic. That is, the probabilistic knowledge-base implicitly contains taxonomic knowledge. Many practical applications in fields like, for example, biology, technology, and medicine require the representation of this kind of taxonomic knowledge besides purely probabilistic knowledge. Own preliminary results in [16] now show that taxonomic knowledge can be exploited for an increased efficiency and a decreased incompleteness in the local approach to probabilistic deduction. Thus, the second motivating idea of this paper is to explore the interplay between taxonomic and probabilistic knowledge in probabilistic deduction, and to elaborate inference rules that exploit taxonomic knowledge. The relationship between taxonomic and probabilistic knowledge is also analyzed in [13], where probabilistic knowledge is integrated into a terminological language.

We choose taxonomic and probabilistic knowledge-bases over conjunctive events as a concrete framework in which our motivating ideas shall be realized. In this framework, the deduction of probabilistic knowledge is NP-hard (we show in [19] that it is even NP-hard for probabilistic knowledge-bases over basic events), while the deduction of taxonomic knowledge can be done in linear time in the size of the taxonomic knowledge-base. Hence, each inference rule that exploits taxonomic knowledge can also be applied in linear time in the size of the taxonomic knowledge-base. Furthermore, taxonomic and probabilistic knowledge-bases over conjunctive events are still expressive enough for many practical applications.

The main contributions of this paper can be summarized as follows. As a first contribution, we present locally complete inference rules for probabilistic deduction from taxonomic and probabilistic knowledge-bases over conjunctive events. More precisely, the presented inference rules deduce logically entailed tightest bounds from a biconnected chain of three conjunctive events under additional taxonomic knowledge over conjunctive events. Crucially, in contrast to existing inference rules in the literature, our inference rules are locally complete for conjunctive events and under additional taxonomic knowledge.

As a second contribution, we discover that the presented inference rules are surprisingly complex and that it is a huge technical effort to work them out and to show their soundness and local completeness. Thus, since it is not obvious at all where the deduced tightest bounds come from, we call them 'magic' inference rules. Hence, it seems unlikely that other locally complete inference rules that have more general or more extensive taxonomic and probabilistic knowledge in their premises can be worked out.

As a third contribution, we show that all systems of inference rules for probabilistic deduction in taxonomic and probabilistic knowledge-bases over conjunctive events are as a matter of principle globally incomplete. In particular, we also provide examples of taxonomic and probabilistic knowledge-bases in which our magic inference rules yield globally very incomplete probabilistic deductions.

The latter contributions are negative results, which are important for the whole probabilistic community. They show that local probabilistic deduction by the iterative application of inference rules on interval restrictions for conditional probabilities is very limited in its field of application.

The rest of this paper is organized as follows. In Section 2, we introduce the technical background of this work. In Section 3, we briefly discuss the computational complexity of probabilistic deduction. Section 4 provides a motivating example. In Sections 5 and 6, we present and discuss our magic inference rules for locally complete probabilistic deduction under taxonomic knowledge. Section 7 summarizes the main results and underlines the general impact of this work.

This paper is a revised extract of own work from [17], which we extended by a short discussion on the computational complexity of probabilistic deduction. Preliminary results of this paper have been presented in [16].

## 2   TECHNICAL PRELIMINARIES

We briefly give a more general introduction to the kind of taxonomic and probabilistic knowledge considered in this work. We deal with taxonomic and probabilistic formulas over propositional events. More precisely, taxonomic formulas represent implications between propositional events, while probabilistic formulas express interval restrictions for conditional probabilities of propositional events. The technical background introduced in this section is commonly accepted in the literature (see, for example, [11] for other work in the same spirit).

We assume a nonempty and finite set of *basic events* $\mathcal{B} = \{B_1, B_2, \ldots, B_n\}$. The set of *conjunctive events* $\mathcal{C}_\mathcal{B}$ comprises the *false event* $\bot$, the *true event* $\top$, and all members in the closure of $\mathcal{B}$ under the Boolean operation $\wedge$. We abbreviate the conjunctive event $C \wedge D$ by $CD$. The set of *propositional events* $\mathcal{G}_\mathcal{B}$ is the closure of $\mathcal{B}$ under the Boolean operations $\wedge$ and $\neg$. We abbreviate the propositional events $B_1 \wedge \neg B_1$ and $\neg(B_1 \wedge \neg B_1)$ by $\bot$ and $\top$, respectively. We abbreviate the propositional events $G \wedge H$ and $\neg G$ by $GH$ and $\overline{G}$, respectively. *Taxonomic formulas* are expressions $G \rightarrow H$ with propositional events $G$ and $H$. *Probabilistic formulas* are expressions $(H|G)[u_1, u_2]$ with real numbers $u_1, u_2 \in [0, 1]$ and propositional events $G$ and $H$. In the probabilistic formula $(H|G)[u_1, u_2]$, we call $G$ the *premise* and $H$ the *conclusion*.

In order to define probabilistic interpretations of propositional events, taxonomic formulas, and probabilistic formulas, we introduce atomic events and the binary relation '$\Rightarrow$' between atomic and propositional events. The set of *atomic events* $\mathcal{A}_{\mathcal{B}}$ is defined by $\mathcal{A}_{\mathcal{B}} = \{E_1 E_2 \cdots E_n \mid E_i = B_i$ or $E_i = \overline{B_i}$ for all $i \in [1:n]\}$. The atomic events of our framework coincide with the more commonly known possible worlds from probabilistic logic [21]. For all atomic events $A$ and all propositional events $G$, let $A \Rightarrow G$ iff $A\overline{G}$ is a propositional contradiction.

A *probabilistic interpretation* $Pr$ is a mapping from $\mathcal{A}_{\mathcal{B}}$ to $[0, 1]$ such that all $Pr(A)$ with $A \in \mathcal{A}_{\mathcal{B}}$ sum up to 1. $Pr$ is extended in a well-defined way to propositional events $G$ by $Pr(G) = \sum_{A \in \mathcal{A}_{\mathcal{B}}, A \Rightarrow G} Pr(A)$. $Pr$ is extended to taxonomic formulas by $Pr \models G \rightarrow H$ iff $Pr(G) = Pr(GH)$. $Pr$ is extended to probabilistic formulas by:

$$Pr \models (H|G)[u_1, u_2] \text{ iff}$$
$$u_1 \cdot Pr(G) \le Pr(GH) \le u_2 \cdot Pr(G).$$

The notions of models, satisfiability, and logical consequence for taxonomic and probabilistic formulas are defined in the classical way. A probabilistic interpretation $Pr$ is a *model* of a formula $F$ iff $Pr \models F$. $Pr$ is a *model* of a set of formulas $KB$, denoted $Pr \models KB$, iff $Pr$ is a model of all $F \in KB$. A set of formulas $KB$ is *satisfiable* iff a model of $KB$ exists. A formula $F$ is a *logical consequence* of a set of formulas $KB$, denoted $KB \models F$, iff each model of $KB$ is also a model of $F$.

For a probabilistic formula $(H|G)[u_1, u_2]$ and a set of taxonomic and probabilistic formulas $KB$, let $u$ denote the set of all real numbers $u \in [0, 1]$ for which there exists a model $Pr$ of $KB$ with $Pr \models (H|G)[u, u]$ and $Pr(G) > 0$. Now, we easily verify that $(H|G)[u_1, u_2]$ is a logical consequence of $KB$ iff $u_1 \le \inf u$ and $u_2 \ge \sup u$.

This observation yields a canonic notion of tightness for logical consequences of probabilistic formulas: the probabilistic formula $(H|G)[u_1, u_2]$ is a *tight logical consequence* of $KB$, denoted $KB \models_{tight} (H|G)[u_1, u_2]$, iff $u_1 = \inf u$ and $u_2 = \sup u$.

Note that $u$ is a closed interval in the real numbers (see, for example, [11] and [17]). For $u = \emptyset$, we canonically define $\inf u = 1$ and $\sup u = 0$. Now, $u = \emptyset$ iff $KB \models (G|\top)[0, 0]$ iff $KB \models_{tight} (H|G)[1, 0]$ iff $KB \models (H|G)[u_1, u_2]$ for all $u_1, u_2 \in [0, 1]$.

Based on the notions of logical consequence and of tight logical consequence, we now define probabilistic deduction problems and their solutions, that is, probabilistic queries to taxonomic and probabilistic knowledge-bases and their correct and tight answers.

A *taxonomic knowledge-base TKB* is a set of taxonomic formulas. A *probabilistic knowledge-base PKB* is a set of probabilistic formulas $(H|G)[u_1, u_2]$ with $u_1 \le u_2$. A *taxonomic and probabilistic knowledge-base KB* is the union of a taxonomic knowledge-base *TKB* and a probabilistic knowledge-base *PKB*. A *probabilistic query* to $KB$ is an expression $\exists (F|E)[x_1, x_2]$ with propositional events $E$ and $F$, and two different variables $x_1$ and $x_2$. Its *tight answer* is the substitution $\sigma = \{x_1/u_1, x_2/u_2\}$ with $u_1, u_2 \in [0, 1]$ such that $KB \models_{tight} (F|E)[u_1, u_2]$. A *correct answer* is a substitution $\sigma = \{x_1/u_1, x_2/u_2\}$ with $u_1, u_2 \in [0, 1]$ such that $KB \models (F|E)[u_1, u_2]$.

Given a probabilistic query $\exists (F|E)[x_1, x_2]$, we consider its tight answer as the desired semantics: first, the tight answer for $\exists (F|E)[x_1, x_2]$ subsumes all correct answers. Second, there is exactly one tight answer for $\exists (F|E)[x_1, x_2]$, while there is generally an infinite number of correct answers. Third, also from the practical point of view, we are interested in the tightest bounds that are entailed by a taxonomic and probabilistic knowledge-base.

Finally, we define the notions of soundness and of completeness related to inference rules and to techniques for probabilistic deduction. An inference rule $KB \vdash F$ is *sound* iff $KB \models F$, where $F$ is a taxonomic or probabilistic formula and $KB$ is a taxonomic and probabilistic knowledge-base. An inference rule $KB \vdash (H|G)[u_1, u_2]$ is sound and *locally complete* iff $KB \models_{tight} (H|G)[u_1, u_2]$. A technique for probabilistic deduction is *sound* iff it computes a correct answer for any given probabilistic query. It is sound and *globally complete* iff it computes the tight answer for any given probabilistic query.

## 3   COMPUTATIONAL COMPLEXITY

In the just introduced framework of taxonomic and probabilistic formulas over propositional events, the problem of computing the tight answer for a probabilistic query is NP-hard, since it is a generalization of the satisfiability problem for probabilistic logic, which is known to be NP-complete from [12]. Moreover, the problem of deciding whether a taxonomic knowledge-base is satisfiable is NP-complete, since it generalizes the NP-complete satisfiability problem for propositional logic, and since it is generalized by the NP-complete satisfiability problem for probabilistic logic. Hence, from the computational complexity point of view, it is reasonable to focus on a more restricted class of probabilistic deduction problems.

Surprisingly, even the problem of computing the tight answer for a probabilistic query over basic events to a probabilistic knowledge-base over basic events is NP-hard, as we show in [19]. While already in the framework of taxonomic formulas over conjunctive events, the problem of deciding whether a taxonomic formula is a logical consequence of a taxonomic knowledge-base can be solved in linear time in the size of the taxonomic knowledge-base. More precisely, taxonomic formulas over conjunctive events are well-known as functional dependencies in database theory (see, for example, [4] and [24]). The results of this area show that deducing taxonomic formulas over conjunctive events from taxonomic knowledge-bases over conjunctive events can be done in linear time by using a hull-operator on the set of all subsets of $\mathcal{B} \cup \{\perp\}$ (see [16] and [17]).

In the sequel, we focus on probabilistic queries over conjunctive events to taxonomic and probabilistic knowledge-bases over conjunctive events. In this framework, the deduction of probabilistic knowledge remains NP-hard. However, at least each inference rule that exploits taxonomic knowledge can be applied in linear time in the size of the taxonomic knowledge-base. The next section provides a medical example, which shows the practical importance of this kind of probabilistic deduction problems.

## 4  EXAMPLE

We consider the following taxonomic knowledge about bacterial infections. Tuberculosis of the lungs (tb) and lepromatous leprosy (lep) are different gram-positive bacterial infections (g-pos). Legionellosis (leg), cholera (chol), and typhoid (typh) are different gram-negative bacterial infections (g-neg). Gram-positive bacterial infections and gram-negative bacterial infections are different bacterial infections ($\top$). The symptoms of tuberculosis are coughing (cough), chest pain (chest), and coughing up blood (cough_bl). The symptoms of leprosy are a stuffy nose (st_nose), and skin lesions and nodules (skin_le_no). This taxonomic knowledge can be expressed by the following taxonomic formulas over $\mathcal{C}_{\mathcal{B}}$ with the set of basic events $\mathcal{B} = \{$tb, lep, g-pos, leg, chol, typh, g-neg, cough, chest, cough_bl, st_nose, skin_le_no$\}$ (note that g-pos $\to \top$ and g-neg $\to \top$ are tautologies):

tb lep $\to$ g-pos, tb lep $\to \perp$, leg $\to$ g-neg,

chol $\to$ g-neg, typh $\to$ g-neg, leg chol $\to \perp$,

leg typh $\to \perp$, chol typh $\to \perp$, g-pos g-neg $\to \perp$,

tb $\to$ cough chest cough_bl, lep $\to$ st_nose skin_le_no.

The symptoms of many diseases cannot be clearly defined, since different human bodies may react in different ways to an infection. We assume the following probabilistic knowledge about the symptoms of legionellosis, cholera, and typhoid. More than 80% of the persons infected by legionellosis have muscle aches (muscle), headache (head), tiredness (tired), dry cough followed by high fever (d_cough_h_fever), and chills (chills). More than 60% have diarrhea (diar). More than 80% of the persons infected by cholera have a mild diarrhea (m_diar). More than 70% of the persons infected by typhoid have relapses (relap). More than 80% have fever (fever), headache, constipation or diarrhea (const_or_diar), rose-colored spots on the trunk (spots), and an enlarged spleen and liver (enl_sp_li). Involving the additional basic events muscle, head, tired, d_cough_h_fever, chills, diar, m_diar, relap, fever, const_or_diar, spots, enl_sp_li and the additional taxonomic formulas

d_cough_h_fever $\to$ cough fever, m_diar $\to$ diar,

diar $\to$ const_or_diar,

we can express this probabilistic knowledge about the symptoms of legionellosis, cholera, and typhoid by the following probabilistic formulas:

(muscle head tired d_cough_h_fever chills|leg)[.8, 1],

(diar|leg)[.6, 1], (m_diar|chol)[.8, 1], (relap|typh)[.7, 1],

(fever head const_or_diar spots enl_sp_li|typh)[.8, 1].

Wondering about the tightest lower and upper bound of the probability that typhoid causes fever and headache, we get the probabilistic query $\exists$(fever head|typh)$[x_1, x_2]$, which yields the tight answer $\sigma = \{x_1/.8, x_2/1\}$.

## 5  THE INFERENCE RULES

The literature contains a variety of different inference rules for deducing probabilistic formulas from probabilistic knowledge-bases. If we analyze all these inference rules more deeply, we make two important observations. First, nearly all the results of local completeness just hold for probabilistic formulas over pairwise different basic events. Second, the interplay between taxonomic and probabilistic knowledge is not fully explored so far. In this section, we now present inference rules that are locally complete for probabilistic formulas over conjunctive events under additional taxonomic knowledge over conjunctive events.

We start with fixing the inference patterns of our inference rules. The premise of all selected inference rules is a taxonomic knowledge-base over conjunctive events and a biconnected chain of three (not necessarily pairwise different) conjunctive events. In detail, it is given by $KB = TKB \cup PKB$, where $TKB$ is an arbitrary taxonomic knowledge-base over conjunctive events and $PKB = \{(B|A)[u_1, u_2], (A|B)[v_1, v_2], (C|B)[x_1, x_2], (B|C)[y_1, y_2]\}$ with conjunctive events $A$, $B$, and $C$. The conclusions of the selected inference rules provide the logically entailed tightest bounds for all probabilistic formulas that can be built from the three conjunctive events $A$, $B$, and $C$. In detail, they are given by (the deduced tightest bounds $z_1$ and $z_2$ are presented at the end of this section):

- SHARPENING: $(B|A)[z_1, z_2]$, $(A|B)[z_1, z_2]$

- CHAINING: $(C|A)[z_1, z_2]$

• FUSION: $(AC|B)[z_1, z_2]$, $(B|AC)[z_1, z_2]$

• COMBINATION: $(C|AB)[z_1, z_2]$, $(AB|C)[z_1, z_2]$

We chose these inference rules, since there is already a quite extensive literature on similar inference rules, which are locally complete for biconnected chains of three pairwise different basic events without any taxonomic knowledge beside (see, for example, [9], [2], [25], [8], and [13]). Hence, the selected inference rules seem to be quite important, and they also have well-explored counterparts in restricted frameworks, which may serve for comparisons.

It remains to compute the deduced tightest bounds in the selected inference rules. Let us first give some examples to get a rough idea on possible problems that may arise to our work. Let $\mathcal{B} = \{A, B, C\}$ and let $KB = TKB \cup PKB$, where $TKB$ is given by Table 1, left side, and $PKB$ is given by the conjunctive events $A$, $B$, and $C$ in Table 1, right side, and by the bounds in Table 2.

Table 1: Taxonomic Knowledge

|     | $TKB$ | $A$ | $B$ | $C$ |
|-----|-------|-----|-----|-----|
| (a) | $\{ABC \to \bot\}$ | A | B | C |
| (b) | $\{C \to A,\ AB \to C\}$ | A | B | C |
| (c) | $\{C \to A\}$ | A | B | C |
| (d) | $\{BC \to A\}$ | A | B | C |
| (e) | $\emptyset$ | A | B | AC |

Table 2: Probabilistic Knowledge

|     | $(B|A)$ | $(A|B)$ | $(C|B)$ | $(B|C)$ |
|-----|---------|---------|---------|---------|
| (a) | $[0.90, 0.95]$ | $[0.90, 0.95]$ | $[0.20, 0.25]$ | $[0.75, 0.80]$ |
| (b) | $[0.85, 0.90]$ | $[0.30, 0.35]$ | $[0.20, 0.25]$ | $[0.75, 0.80]$ |
| (c) | $[0.90, 0.95]$ | $[0.30, 0.35]$ | $[0.20, 0.25]$ | $[0.75, 0.80]$ |
| (d) | $[0.90, 0.95]$ | $[0.30, 0.35]$ | $[0.20, 0.25]$ | $[0.75, 0.80]$ |
| (e) | $[0.90, 0.95]$ | $[0.30, 0.35]$ | $[0.20, 0.25]$ | $[0.75, 0.80]$ |

At first sight, the examples (a) to (e) seem harmless. However, $KB$ in (b), (c), and (e) logically entails $A \to \bot$ and $C \to \bot$. Moreover, $KB$ in (a), (c), (d), and (e) logically entails $B \to \bot$. Hence, each probabilistic knowledge-base in (a) to (e) contains at least one probabilistic formula with a false premise. Of course, we should exclude taxonomic and probabilistic knowledge-bases like the ones in (a) to (e) from the premises of our inference rules:

A taxonomic and probabilistic knowledge-base $KB$ is *inconsistent* iff it contains at least one $(B|A)[u_1, u_2]$ with $KB \models A \to \bot$. A taxonomic and probabilistic knowledge-base $KB$ is *consistent* iff it is not inconsistent.

Where do the false premises in the probabilistic formulas of $PKB$ come from? Interestingly, $KB$ is always consistent if we assume that $TKB = \emptyset$ and that $A$, $B$, and $C$ are three pairwise different basic events. However, an inconsistency may arise if we have explicit taxonomic knowledge in

$TKB$ or implicit taxonomic knowledge in the structure of the conjunctive events $A$, $B$, and $C$ (for example, if $A = \mathsf{A}$ and $C = \mathsf{AC}$, then $\emptyset \models C \to A$, and thus $TKB \models C \to A$ for all taxonomic knowledge-bases $TKB$).

The next theorem characterizes the consistency of the premises of our inference rules. It requires the following notion of coherence: $KB = TKB \cup PKB$ is *coherent* iff for all $(B|A)[u_1, u_2] \in PKB$: $TKB \models AB \to \bot \Leftrightarrow u_2 = 0$, and $TKB \models A \to B \Leftrightarrow u_1 = 1$.

**Theorem 5.1** *Let $KB = TKB \cup PKB$ be a coherent taxonomic and probabilistic knowledge-base, where $TKB$ is an arbitrary taxonomic knowledge-base and $PKB = \{(B|A)[u_1, u_2], (A|B)[v_1, v_2], (C|B)[x_1, x_2], (B|C)[y_1, y_2]\}$.*

*$KB$ is inconsistent iff one of the conditions (1) to (7) holds. If one of (1) to (4) holds, then $KB \models A \to \bot$, $C \to \bot$. If one of (3) to (7) holds, then $KB \models B \to \bot$.*

*(1) $TKB \models A \to C$, $BC \to A$ and $u_2 < y_1$,*

*(2) $TKB \models C \to A$, $AB \to C$ and $u_1 > y_2$,*

*(3) $TKB \models A \to C$ and $u_2 x_2 (1 - y_1) < v_1 y_1 (1 - u_2)$,*

*(4) $TKB \models C \to A$ and $u_1 x_1 (1 - y_2) > v_2 y_2 (1 - u_1)$,*

*(5) $TKB \models AB \to C$ and $v_1 > x_2$,*

*(6) $TKB \models BC \to A$ and $v_2 < x_1$,*

*(7) $TKB \models ABC \to \bot$ and $x_1 + v_1 > 1$.*

**Proof.** The proof is given in full detail in [17]. □

Coming back to our examples, for (a) with $TKB = \{ABC \to \bot\}$, we get $TKB \models ABC \to \bot$ and $x_1 + v_1 = 0.2 + 0.9 = 1.1 > 1$. Hence, by Theorem 5.1, $KB$ is inconsistent with $KB \models B \to \bot$. For (e) with $A = \mathsf{A}$ and $C = \mathsf{AC}$, we get $\emptyset \models C \to A$. Thus, $TKB \models C \to A$ and $u_1 x_1 (1 - y_2) = 0.9 \cdot 0.2 \cdot (1 - 0.8) > 0.35 \cdot 0.8 \cdot (1 - 0.9) = v_2 y_2 (1 - u_1)$. Hence, by Theorem 5.1, $KB$ is inconsistent with $KB \models A \to \bot$, $B \to \bot$, $C \to \bot$.

In summary, the premises of our inference rules must be coherent and consistent. The coherence can be checked by simply applying its plain definition, while the consistency can thereafter be checked with Theorem 5.1.

We are ready to proceed with our inference rules. Again, before focusing on their technical details, let us give some illustrating examples. Let $\mathcal{B} = \{A, B, C\}$ and let $KB = TKB \cup PKB$, where $TKB$ is given by Table 3, left side, and $PKB$ is given by the conjunctive events $A$, $B$, and $C$ in Table 3, right side, and by the bounds in Table 4. We easily verify that all $KB$ in (f) to (k) are coherent and consistent.

Tables 5 to 7 show the tight logical consequences of $KB$ that are deducible by our inference rules SHARPENING, CHAINING, FUSION, and COMBINATION (the underlined bounds for SHARPENING improve the given bounds).

Table 3: Taxonomic Knowledge

|     | TKB | A | B | C |
|-----|-----|---|---|---|
| (f) | $\{ABC \to \bot\}$ | A | B | C |
| (g) | $\{C \to A, AB \to C\}$ | A | B | C |
| (h) | $\{C \to A\}$ | A | B | C |
| (i) | $\{BC \to A\}$ | A | B | C |
| (j) | $\emptyset$ | A | B | AC |
| (k) | $\emptyset$ | A | B | C |

Table 4: Probabilistic Knowledge

|     | $(B\|A)$ | $(A\|B)$ | $(C\|B)$ | $(B\|C)$ |
|-----|----------|----------|----------|----------|
| (f) | $[0.90, 0.95]$ | $[0.10, 0.15]$ | $[0.20, 0.25]$ | $[0.75, 0.80]$ |
| (g) | $[0.60, 0.65]$ | $[0.30, 0.35]$ | $[0.25, 0.30]$ | $[0.75, 0.80]$ |
| (h) | $[0.85, 0.90]$ | $[0.30, 0.35]$ | $[0.20, 0.25]$ | $[0.75, 0.80]$ |
| (i) | $[0.90, 0.95]$ | $[0.30, 0.35]$ | $[0.20, 0.25]$ | $[0.75, 0.80]$ |
| (j) | $[0.85, 0.90]$ | $[0.30, 0.35]$ | $[0.20, 0.25]$ | $[0.75, 0.80]$ |
| (k) | $[0.85, 0.90]$ | $[0.30, 0.35]$ | $[0.20, 0.25]$ | $[0.75, 0.80]$ |

Table 5: SHARPENING

|     | $(B\|A)$ | $(A\|B)$ | $(C\|B)$ | $(B\|C)$ |
|-----|----------|----------|----------|----------|
| (f) | $[0.90, 0.95]$ | $[0.10, 0.15]$ | $[0.20, 0.25]$ | $[0.75, 0.80]$ |
| (g) | $[0.60, 0.65]$ | $[0.30, \underline{0.30}]$ | $[\underline{0.30}, 0.30]$ | $[0.75, 0.80]$ |
| (h) | $[0.85, \underline{0.88}]$ | $[0.30, 0.35]$ | $[0.20, 0.25]$ | $[\underline{0.76}, 0.80]$ |
| (i) | $[0.90, 0.95]$ | $[0.30, 0.35]$ | $[0.20, 0.25]$ | $[0.75, 0.80]$ |
| (j) | $[0.85, \underline{0.88}]$ | $[0.30, 0.35]$ | $[0.20, 0.25]$ | $[\underline{0.76}, 0.80]$ |
| (k) | $[0.85, 0.90]$ | $[0.30, 0.35]$ | $[0.20, 0.25]$ | $[0.75, 0.80]$ |

Table 6: CHAINING and FUSION

|     | $(C\|A)$ | $(A\|C)$ | $(B\|AC)$ | $(AC\|B)$ |
|-----|----------|----------|-----------|-----------|
| (f) | $[0.00, 0.10]$ | $[0.00, 0.07]$ | $[0.00, 0.00]$ | $[0.00, 0.00]$ |
| (g) | $[0.75, 0.87]$ | $[1.00, 1.00]$ | $[0.75, 0.80]$ | $[0.30, 0.30]$ |
| (h) | $[0.61, 0.75]$ | $[1.00, 1.00]$ | $[0.76, 0.80]$ | $[0.20, 0.25]$ |
| (i) | $[0.51, 0.85]$ | $[0.75, 0.96]$ | $[0.84, 1.00]$ | $[0.20, 0.25]$ |
| (j) | $[0.61, 0.75]$ | $[1.00, 1.00]$ | $[0.76, 0.80]$ | $[0.20, 0.25]$ |
| (k) | $[0.00, 0.86]$ | $[0.00, 1.00]$ | $[0.00, 1.00]$ | $[0.20, 0.25]$ |

Table 7: COMBINATION

|     | $(C\|AB)$ | $(AB\|C)$ | $(A\|BC)$ | $(BC\|A)$ |
|-----|-----------|-----------|-----------|-----------|
| (f) | $[0.00, 0.00]$ | $[0.00, 0.00]$ | $[0.00, 0.00]$ | $[0.00, 0.00]$ |
| (g) | $[1.00, 1.00]$ | $[0.75, 0.80]$ | $[1.00, 1.00]$ | $[0.60, 0.65]$ |
| (h) | $[0.57, 0.71]$ | $[0.76, 0.80]$ | $[1.00, 1.00]$ | $[0.49, 0.60]$ |
| (i) | $[0.57, 0.83]$ | $[0.75, 0.80]$ | $[1.00, 1.00]$ | $[0.51, 0.79]$ |
| (j) | $[0.57, 0.71]$ | $[0.76, 0.80]$ | $[1.00, 1.00]$ | $[0.49, 0.60]$ |
| (k) | $[0.00, 0.83]$ | $[0.00, 0.80]$ | $[1.00, 1.00]$ | $[0.00, 0.75]$ |

The examples (f) to (i) contain explicit taxonomic knowledge in the taxonomic knowledge-base, while the example (j) contains implicit taxonomic knowledge in the structure of the conjunctive events ($A = $ A and $C = $ AC entails $\emptyset \models C \to A$, hence $TKB \models C \to A$).

We observe that the deduced tightest bounds in the examples with explicit or implicit taxonomic knowledge are much tighter than the ones in the examples without any taxonomic knowledge at all: the examples (h) and (j) increase (k) by exactly the additional explicit and implicit, respectively, taxonomic knowledge $C \to A$. As a consequence, the deduced tightest bounds in (h) and (j) are much tighter than the ones in (k). For instance, CHAINING deduces $(C|A)[0.61, 0.75]$ in (h) and (j) compared to only $(C|A)[0.00, 0.86]$ in (k), and FUSION deduces $(B|AC)[0.76, 0.80]$ in (h) and (j) compared to only $(B|AC)[0.00, 1.00]$ in (k).

The fact that implicit taxonomic knowledge may increase the tightness of the deduced bounds also shows that all similar inference rules of the literature that are locally complete for a biconnected chain of three pairwise different basic events are generally not locally complete anymore for a biconnected chain of three conjunctive events.

Finally, we present our *magic inference rules*:

**Theorem 5.2** *Let $KB = TKB \cup PKB$ be a coherent and consistent taxonomic and probabilistic knowledge-base, where $TKB$ is an arbitrary taxonomic knowledge-base and $PKB = \{(B|A)[u_1, u_2], (A|B)[v_1, v_2], (C|B)[x_1, x_2], (B|C)[y_1, y_2]\}$. In the sequel, we abbreviate $TKB \models ABC \to \bot$ by $\alpha$, $TKB \models C \to A$ by $\beta$, $TKB \models A \to C$ by $\gamma$, $TKB \models BC \to A$ by $\delta$, $TKB \models AB \to C$ by $\varepsilon$, and $TKB \models AC \to B$ by $\zeta$.*

*The operands of min and max may be followed by a set of conditions that must all hold for including the operand in computing the minimum and maximum, respectively (for example, $\min(v_2, x_2, y_2\{\beta, \varepsilon\})$ denotes $\min(v_2, x_2, y_2)$ if both $\beta$ and $\varepsilon$ hold, and $\min(v_2, x_2)$ otherwise).*

SHARPENING:

*a) $KB \models_{tight} (B|A)[z_1, z_2]$ with*

$$z_1 = \max(u_1, \tfrac{v_1 y_1}{v_1 y_1 + x_2(1 - y_1)} \{\gamma, v_1 y_1 > 0\}, y_1 \{\gamma, \delta\})$$

$$z_2 = \min(u_2, \tfrac{v_2 y_2}{v_2 y_2 + x_1(1 - y_2)} \{\beta, v_2 y_2 > 0\}, y_2 \{\beta, \varepsilon\}).$$

*b) $KB \models_{tight} (A|B)[z_1, z_2]$ with*

$$z_1 = \max(\tfrac{u_1 x_1 (1 - y_2)}{y_2 (1 - u_1)} \{\beta, 1 > u_1 > y_2 > 0\},$$
$$v_1, x_1 \{\delta\})$$

$$z_2 = \min(1 - x_1 \{\alpha\}, \tfrac{u_2 x_2 (1 - y_1)}{y_1 (1 - u_2)} \{\gamma, y_1 > u_2\},$$
$$v_2, x_2 \{\varepsilon\}).$$

CHAINING:

$KB \models_{tight} (C|A)[z_1, z_2]$ with

$$z_1 = \max(0, u_1 + \tfrac{u_1}{v_1} + \tfrac{u_1 x_1}{v_1} \{v_1 + x_1 > 1\}, u_1 \{\varepsilon\},$$
$$\tfrac{u_1 x_1}{v_2} \{\delta, v_2 > 0\}, \tfrac{u_1 x_1}{v_2 y_2} \{\beta, v_2 y_2 > 0\},$$
$$\tfrac{u_1}{y_2} \{\beta, \varepsilon, y_2 > 0\}, 1 \{\gamma\})$$

$$z_2 = \min(1, 1 - u_1 + \tfrac{u_1 x_2}{v_1}\{v_1 > x_2\}, \tfrac{u_2 x_2}{v_1 y_1}\{v_1 y_1 > 0\},$$
$$\tfrac{x_2}{v_1 y_1 + x_2(1-y_1)}\{v_1 > x_2, y_1 > 0\}, 1 - u_1\{\alpha\},$$
$$u_2 - \tfrac{u_2 x_2}{v_1} + \tfrac{u_2 x_2}{v_1 y_1}\{v_1 y_1 > 0\}, \tfrac{u_2}{v_1}\{\delta, y_1 > u_2\},$$
$$\tfrac{1-u_1}{1-y_2}\{\beta, u_1 > y_2\}, \tfrac{u_2 x_2}{v_1}\{\zeta, v_1 > x_2\}, u_2\{\zeta\},$$
$$\tfrac{u_2(1-y_1)\min(x_2, 1-v_1)}{v_1 y_1}\{\alpha, v_1 y_1 > 0\}, 0\{\alpha, \zeta\},$$
$$\tfrac{(1-y_1)\min(x_2, 1-v_1)}{v_1 y_1 + (1-y_1)\min(x_2, 1-v_1)}\{\alpha, v_1 y_1 > 0\}).$$

FUSION:

a) *If* $TKB \not\models AC \to \bot$, *then*

$KB \models_{tight} (B|AC)[z_1, z_2]$ *with*

$$z_1 = \max(\max(\tfrac{y_1(v_1 + x_1 - 1)}{y_1(v_1 - 1) + x_1}, \tfrac{u_1(x_1 + v_1 - 1)}{u_1(x_1 - 1) + v_1})\{x_1 + v_1 > 1\},$$
$$u_1\{\varepsilon\}, y_1\{\delta\}, \tfrac{v_1 y_1}{v_1 y_1 + x_2(1-y_1)}\{\varepsilon, v_1 y_1 > 0\},$$
$$\tfrac{x_1 u_1}{x_1 u_1 + v_2(1-u_1)}\{\delta, x_1 u_1 > 0\}, 0, 1\{\zeta\})$$
$$z_2 = \min(1, u_2\{\gamma\}, y_2\{\beta\}, 0\{\alpha\}).$$

b) $KB \models_{tight} (AC|B)[z_1, z_2]$ *with*

$$z_1 = \max(0, x_1 + v_1 - 1, x_1\{\delta\}, v_1\{\varepsilon\})$$
$$z_2 = \min(v_2, x_2, \tfrac{u_2 x_2(1-y_1)}{y_1(1-u_2)}\{\gamma, y_1 > u_2\},$$
$$\tfrac{v_2 y_2(1-u_1)}{u_1(1-y_2)}\{\beta, u_1 > y_2\}, 0\{\alpha\}).$$

COMBINATION:

a) *If* $TKB \not\models AB \to \bot$, *then*

$KB \models_{tight} (C|AB)[z_1, z_2]$ *with*

$$z_1 = \max(0, 1 - \tfrac{1}{v_1} + \tfrac{x_1}{v_1}\{v_1 + x_1 > 1\}, 1\{\varepsilon\},$$
$$\tfrac{x_1}{v_2}\{\delta, v_2 > 0\})$$
$$z_2 = \min(1, \tfrac{y_2(1-u_1)}{u_1(1-y_2)}\{\beta, u_1 > y_2\}, 0\{\alpha\},$$
$$\tfrac{x_2}{v_1}\{v_1 > x_2\})$$

b) $KB \models_{tight} (AB|C)[z_1, z_2]$ *with*

$$z_1 = \max(0, \tfrac{v_1 y_1}{x_1} - \tfrac{y_1}{x_1} + y_1\{v_1 + x_1 > 1\}, y_1\{\delta\},$$
$$u_1\{\beta, \varepsilon\}, \tfrac{u_1 x_1}{u_1 x_1 + v_2(1-u_1)}\{\beta, u_1 x_1 > 0\},$$
$$\tfrac{v_1 y_1}{x_2}\{\varepsilon, x_2 > 0\})$$
$$z_2 = \min(\tfrac{v_2 y_2}{x_1}\{x_1 > v_2\}, \tfrac{u_2(1-y_1)}{1-u_2}\{\gamma, 1 > u_2\},$$
$$\tfrac{u_2 v_2}{u_2 x_1 + v_2(1-u_2)}\{\gamma, x_1 > v_2 > 0\}, 0\{\alpha\},$$
$$y_2, u_2\{\gamma\}).$$

**Proof.** The proof is given in full detail in [17]. □

# 6 DISCUSSION

In the previous section, we presented the magic inference rules SHARPENING, CHAINING, FUSION, and COMBINATION, which deduce tight logical consequences from a biconnected chain of three conjunctive events under additional taxonomic knowledge over conjunctive events.

We discover that our magic inference rules are surprisingly complex. At first glance, it is not clear at all where the deduced tightest bounds come from (this is the reason for which we call them 'magic' inference rules). In [17], we need a huge technical effort to discover these bounds, and to prove soundness and local completeness of the magic inference rules. Hence, it seems unlikely that other locally complete inference rules that have more extensive taxonomic and probabilistic knowledge in their premises can be worked out. Also, just generalizing our inference rules to propositional events would be a nearly intractable task.

Another interesting result is revealed if we analyze the global completeness of a probabilistic deduction technique that is based on the iterative application of the magic inference rules. Since we put a huge effort in elaborating our locally complete magic inference rules, we may at least hope that they are also a big step towards global completeness. However, we now show that all systems of inference rules for probabilistic deduction in taxonomic and probabilistic knowledge-bases over conjunctive events are globally incomplete (note that we assume a fixed number of probabilistic formulas in the premise of each inference rule).

We give an indirect proof of this important result: let us assume that we have a globally complete system of inference rules in which the number of probabilistic formulas in the premise of each inference rule is limited by $k \geq 1$. Now, let $\mathcal{B} = \{B_1, B_2, \ldots, B_n\}$ with $n \geq k + 2$ and let $KB = TKB \cup PKB$ be given by $TKB = \{B_i B_j \to \bot \mid 1 \leq i < j \leq n\}$ and $PKB = \{(B_i|\top)[1/n, 1] \mid 1 \leq i \leq n\}$. We get the tight logical consequence $KB \models_{tight} (B_1|\top)[1/n, 1/n]$. However, the least upper bound $1/n$ cannot be deduced by the assumed system of inference rules, since it requires all the lower bounds of the $n - 1 > k$ probabilistic formulas $(B_i|\top)[1/n, 1]$ with $i \in [2{:}n]$. We also cannot divide the computation, since we do not have any probabilistic formulas over conjunctive events that could keep provisional results. Note, however, that with probabilistic formulas over propositional events, the computation could be divided: for example, for $n = k + 2$, we could deduce first $(B_2 \lor B_3|\top)[2/n, 2/n]$ (that is, $(\neg(\neg B_2 \land \neg B_3)|\top)[2/n, 2/n]$) and thereafter $(B_1|\top)[1/n, 1/n]$, assuming an appropriate system of inference rules.

Hence, also our magic inference rules are globally incomplete, since the maximum number of probabilistic formulas in their premises is four. In the considered example, our magic inference rules deduce the upper bound $1 - 1/n$, which is different from the least upper bound $1/n$ already for $n > 2$. Taking, for example, $n = 100$, the deduced upper bound is 0.99, but the least upper bound is 0.01.

We give another example, which shows that the iterative application of CHAINING may globally be very incomplete. Let $\mathcal{B} = \{B_1, B_2, B_3, B_4\}$, $TKB = \emptyset$, and $PKB = \bigcup \{\{(D|C)[0.1, 0.15], (C|D)[0.8, 1]\} \mid (C, D) \in$ $\{(B_1, B_2), (B_2, B_3), (B_3, B_4)\}\}$. We get the tight logical consequence $TKB \cup PKB \models_{tight} (B_4|B_1)[0, 0.007]$. However, the iterative application of CHAINING just deduces the interval $[0, 0.904]$. Note that we analyze more general probabilistic deduction problems with probabilistic formulas over basic events in [19].

In summary, there is a huge effort in exploring the 'magic' of locally complete inference rules for probabilistic deduction from taxonomic and probabilistic knowledge-bases over conjunctive events. Moreover, as a matter of principle, there does not exist any globally complete system of inference rules for this framework.

# 7   SUMMARY AND CONCLUSIONS

We presented locally complete inference rules for probabilistic deduction from taxonomic and probabilistic knowledge-bases over conjunctive events. Surprisingly, these inference rules are very complex and it is at first glance not clear at all where the deduced tightest bounds come from. Moreover, analyzing the global completeness of our inference rules, we discovered examples of globally very incomplete probabilistic deductions. More generally, we even showed that all systems of inference rules for taxonomic and probabilistic knowledge-bases over conjunctive events are globally incomplete.

Hence, probabilistic deduction by the iterative application of inference rules on probabilistic formulas seems very limited in its field of application. The way in which probabilistic interpretations give semantics to probabilistic formulas seems to contradict the kind of modularity that stands behind the iterative application of inference rules. This important insight has an impact on all areas that deal with probabilistic deduction in similar frameworks.

## References

[1] E. W. Adams. *The Logic of Conditionals*, volume 86 of *Synthese Library*. D. Reidel, Dordrecht, Holland, 1975.

[2] S. Amarger, D. Dubois, and H. Prade. Constraint propagation with imprecise conditional probabilities. In *Proc. of the 7th Conference on Uncertainty in Artificial Intelligence*, pages 26–34. Morgan Kaufmann Publishers, 1991.

[3] K. A. Andersen and J. N. Hooker. Bayesian logic. *Decision Support Systems*, 11:191–210, 1994.

[4] C. Beeri and P. A. Bernstein. Computational problems related to the design of normal form relational schemas. *ACM Transactions on Database Systems*, 4:30–59, 1979.

[5] R. Carnap. *Logical Foundations of Probability*. University of Chicago Press, Chicago, 1950.

[6] B. de Finetti. *Theory of Probability*. Wiley, New York, 1974.

[7] D. Dubois and H. Prade. On fuzzy syllogisms. *Computational Intelligence*, 4(2):171–179, 1988.

[8] D. Dubois, H. Prade, L. Godo, and R. L. de Màntaras. A symbolic approach to reasoning with linguistic quantifiers. In *Proc. of the 8th Conf. on Uncertainty in Artificial Intelligence*, pages 74–82. Morgan Kaufmann Publishers, 1992.

[9] D. Dubois, H. Prade, and J.-M. Touscas. Inference with imprecise numerical quantifiers. In *Intelligent Systems*, chapter 3, pages 53–72. Ellis Horwood, 1990.

[10] R. Fagin, J. Y. Halpern, and N. Megiddo. A logic for reasoning about probabilities. *Information and Computation*, 87:78–128, 1990.

[11] A. M. Frisch and P. Haddawy. Anytime deduction for probabilistic logic. *Artificial Intelligence*, 69:93–122, 1994.

[12] G. Georgakopoulos, D. Kavvadias, and C. H. Papadimitriou. Probabilistic satisfiability. *J. of Complexity*, 4:1–11, 1988.

[13] J. Heinsohn. Probabilistic description logics. In *Proc. of the 10th Conference on Uncertainty in Artificial Intelligence*, pages 311–318. Morgan Kaufmann Publishers, 1994.

[14] B. Jaumard, P. Hansen, and M. P. de Aragão. Column generation methods for probabilistic logic. *ORSA Journal of Computing*, 3:135–147, 1991.

[15] D. Kavvadias and C. H. Papadimitriou. A linear programming approach to reasoning about probabilities. *Annals of Mathematics and Artificial Intelligence*, 1:189–205, 1990.

[16] T. Lukasiewicz. Uncertain reasoning in concept lattices. In *Proc. of the 3rd European Conference on Symbolic and Quantitative Approaches to Reasoning and Uncertainty*, volume 946 of *LNCS/LNAI*, pages 293–300. Springer, 1995.

[17] T. Lukasiewicz. *Precision of Probabilistic Deduction under Taxonomic Knowledge*. Doctoral Dissertation, Universität Augsburg, 1996.

[18] T. Lukasiewicz. Efficient global probabilistic deduction from taxonomic and probabilistic knowledge-bases over conjunctive events. In *Proc. of the 6th International Conference on Information and Knowledge Management*, pages 75–82. ACM Press, 1997.

[19] T. Lukasiewicz. Probabilistic deduction with conditional constraints over basic events. In *Principles of Knowledge Representation and Reasoning: Proc. of the 6th International Conference*. Morgan Kaufmann Publishers, 1998.

[20] T. Lukasiewicz. Probabilistic logic programming. In *Proc. of the 13th European Conference on Artificial Intelligence*, pages 388–392. J. Wiley & Sons, 1998. To appear.

[21] N. J. Nilsson. Probabilistic logic. *Artificial Intelligence*, 28:71–88, 1986.

[22] G. Paaß. Probabilistic Logic. In *Non-Standard Logics for Automated Reasoning*, chapter 8, pages 213–251. Academic Press, London, 1988.

[23] J. Pearl. *Probabilistic Reasoning in Intelligent Systems: Networks of Plausible Inference*. Morgan Kaufmann Publishers, San Mateo, California, 1988.

[24] Y. Sagiv, C. Delobel, D. S. Parker, and R. Fagin. An equivalence between relational database dependencies and a fragment of propositional logic. *Journal of the ACM*, 28(3):435–453, 1981.

[25] H. Thöne, U. Güntzer, and W. Kießling. Towards precision of probabilistic bounds propagation. In *Proc. of the 8th Conference on Uncertainty in Artificial Intelligence*, pages 315–322. Morgan Kaufmann Publishers, 1992.